\documentclass[sigconf]{aamas} 

\usepackage{balance} 
\usepackage{soul}
\usepackage{subfigure}
\usepackage{amsmath}
\usepackage{mathtools}
\usepackage{amsthm}

\newtheorem{definition}{Definition}

\setcopyright{ifaamas}
\acmConference[AAMAS '23]{Proc.\@ of the 22nd International Conference
on Autonomous Agents and Multiagent Systems (AAMAS 2023)}{May 29 -- June 2, 2023}
{London, United Kingdom}{A.~Ricci, W.~Yeoh, N.~Agmon, B.~An (eds.)}
\copyrightyear{2023}
\acmYear{2023}
\acmDOI{}
\acmPrice{}
\acmISBN{}

\acmSubmissionID{???}

\title[AAMAS-2023 Formatting Instructions]{Causality Detection for Efficient Multi-Agent Reinforcement Learning}
\subtitle{Extended Abstract}

\author{Rafael Pina}
\affiliation{
  \institution{Loughborough University London}
  \city{London}
  \country{United Kingdom}}
\email{r.m.pina@lboro.ac.uk}

\author{Varuna De Silva}
\affiliation{
  \institution{Loughborough University London}
  \city{London}
  \country{United Kingdom}}
\email{v.d.de-silva@lboro.ac.uk}

\author{Corentin Artaud}
\affiliation{
  \institution{Loughborough University London}
  \city{London}
  \country{United Kingdom}}
\email{c.artaud2@lboro.ac.uk}

\begin{abstract}
When learning a task as a team, some agents in Multi-Agent Reinforcement Learning (MARL) may fail to understand their true impact in the performance of the team. Such agents end up learning sub-optimal policies, demonstrating undesired lazy behaviours. To investigate this problem, we start by formalising the use of temporal causality applied to MARL problems. We then show how causality can be used to penalise such lazy agents and improve their behaviours. By understanding how their local observations are causally related to the team reward, each agent in the team can adjust their individual credit based on whether they helped to cause the reward or not. We show empirically that using causality estimations in MARL improves not only the holistic performance of the team, but also the individual capabilities of each agent. We observe that the improvements are consistent in a set of different environments. 
\end{abstract}


\newcommand{\BibTeX}{\rm B\kern-.05em{\sc i\kern-.025em b}\kern-.08em\TeX}

\begin{document}


\pagestyle{fancy}
\fancyhead{}


\maketitle 


\section{Introduction}\label{sec:intro}
In cooperative Multi-Agent Reinforcement Learning (MARL) multiple agents must learn strategies as a team in order to solve certain tasks where their cooperation is needed. In this type of problems, each agent is required to interact with the teammates in such a way that leads the team to achieve a common goal \cite{gupta_cooperative_2017}. 

Centralised training decentralised execution (CTDE) \cite{olienoek_2008,KRAEMER201682} is a paradigm adopted by several popular approaches in MARL. In this paradigm, during execution the actions of the agents are only conditioned on their local observations, but during training they have access to extra state information of the environment. This creates a reasonable balance between two inherent problems in MARL: the exponential growth of the action space and the non-stationarity \cite{de_witt_multi-agent_2020}. However, one problem that still remains is known as the lazy agent pathology and can happen when the credit assignment to the agents is not correct \cite{sunehag_value-decomposition_2018}. As the name suggests, lazy agents are agents that do not cooperate towards the common overall goal of the team due to learning sub-optimal policies. However, in MARL the reward is given to the whole team, leading these agents to think that they always contribute to the reward received.

With this work, we aim to bridge the concepts of temporal causality and MARL. By introducing Independent Causal Learning (ICL), we demonstrate that, if we can detect causal relations between individual observations and rewards in MARL, then we can address the lazy agent problem. Furthermore, we demonstrate how independent agents develop more cooperative and more intelligent behaviours when compared to simple fully independent learners. Additionally, we show that causal relations in multi-agent systems between individual observations of agents and the team reward can be detected by Amortized Causal Discovery (ACD) \cite{lowe_amortized_2022}, a state-of-the-art causality detection method. 

\section{Methods}\label{sec:meth}
\subsection{Independent Causal Learning (ICL)}\label{sec:icmarl}
In this section we present Independent Causal Learning (ICL). The goal of this method is to show that causality detection can be used in cooperative tasks by punishing lazy agents. At the same time, we create a ground truth for the next method ACD-MARL, that uses ACD. To formalise the proposed method, we build the link between Granger Causality \cite{granger_1969} and MARL scenarios based on the definitions of Granger Causality for non-linear systems \cite{Tank_2021,lowe_amortized_2022}. Thus, we can write the definition:
\begin{definition}\label{def:te_2}
Let $E$ represent a certain episode sampled from a replay buffer of experiences in MARL, denoting a set of time series of $N$ observations and rewards $E=\{o_1,\dots,o_N,r\}$. Given a non-linear function $g_r$ that maps a set of past values to the series $r$, for the set $E$ we can say that a series $o_i$ Granger-causes the series $r$ if $g_r$ depends on the past values of $o_i$. Formally, we can write $\exists o_i^{'\leq t} \neq o_i^{\leq t} : g_r(o_1^{\leq t},\dots, o_i^{'\leq t},\dots, o_N^{\leq t},r) \neq g_r(o_1^{\leq t},\dots, o_i^{\leq t},\dots, o_N^{\leq t},r)$.
\end{definition} 

This definition bridges the concepts of causality and MARL. We can now define a rule for the calculation of each individual causality factor $c_i$, and how they are used in the learning problem,
\begin{equation}\label{eq:theo_eq1}
c_i(o_i,r)=\left\{
\begin{array}{ll}
    1 & o_i\ causes\ r\\
    0 & \lnot\ o_i\ causes\ r
\end{array}, i \in \{1,\ldots,N\}
\right.    
\end{equation}
and, as a result, we can adjust the Q-update as it follows:
\begin{equation}\label{eq:theo_eq2}
    Q_i(\tau_i,a_i)=(1-\alpha)Q_i(\tau_i,a_i)+\alpha\left[c_i(\tau_i,r)\times r+\gamma\mathop{\mathrm{max}}_{a_i'}Q_i(\tau_i',a_i')\right]
\end{equation}

\begin{figure}[!t]
    \centering
    \includegraphics[width=0.6\columnwidth]{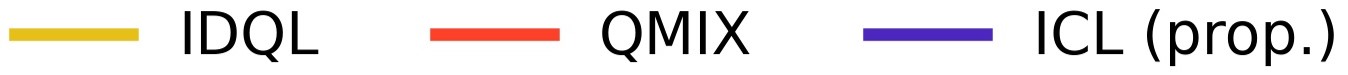}
    \\
    \vspace{0.00mm} 
    \subfigure[Predator-Prey]{\label{fig:res_a}\includegraphics[width=0.45\columnwidth]{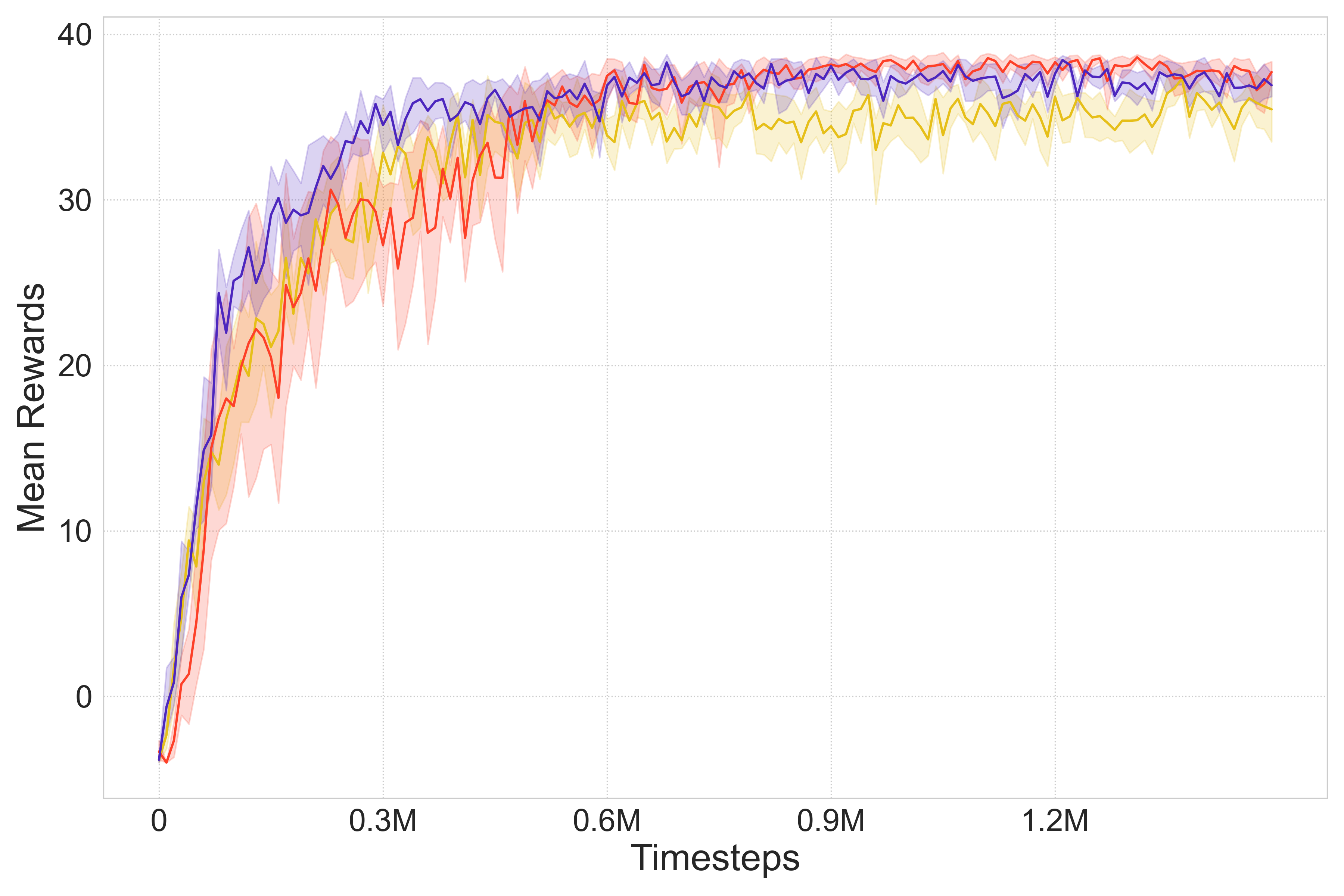}}
    \subfigure[Lumberjacks]{\label{fig:res_c}\includegraphics[width=0.45\columnwidth]{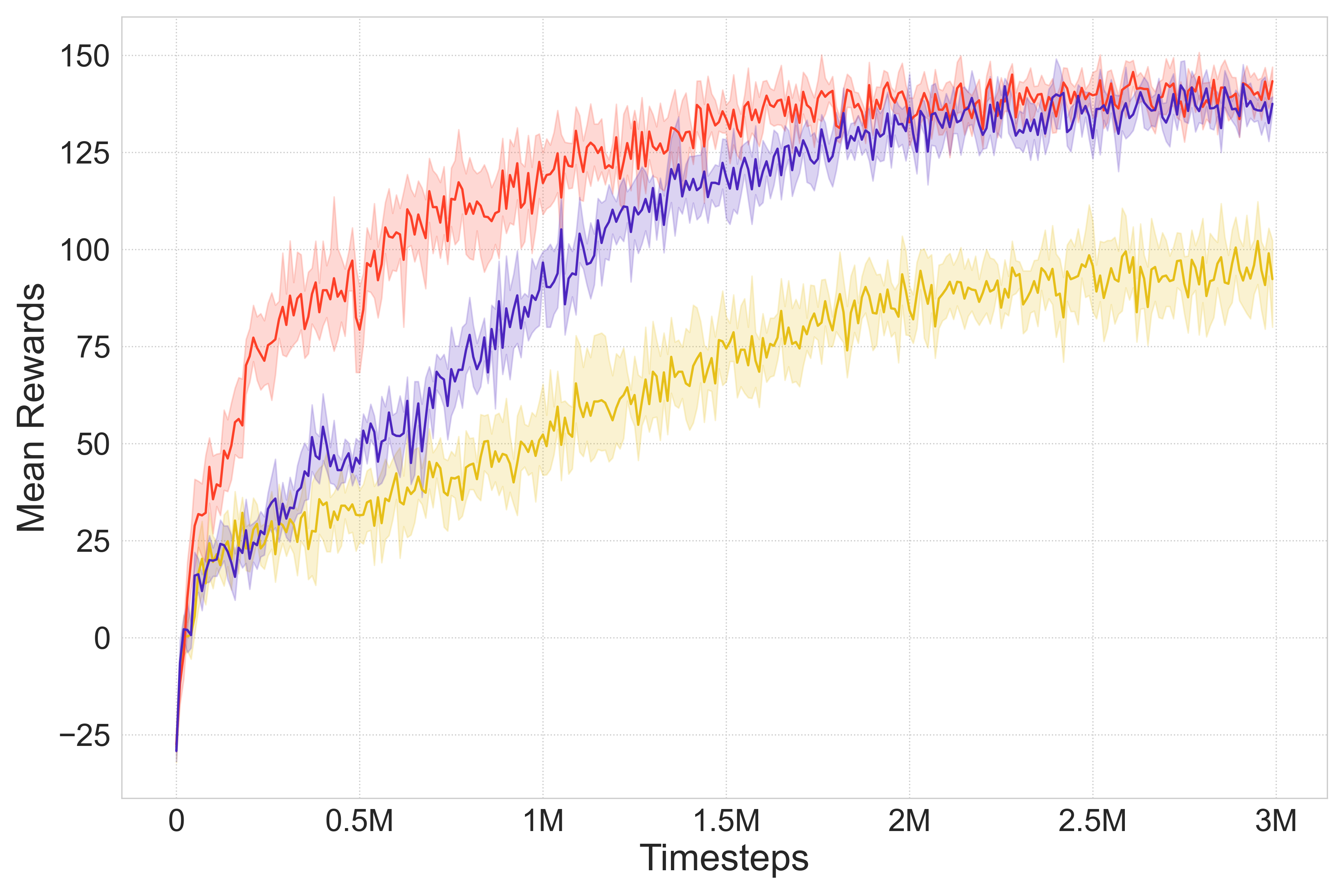}}
    \caption{Team rewards achieved in the environments.}
    \label{fig:results}
\end{figure}

To let agents learn independently, in this work we consider independent policy networks $\pi_i$ (except for QMIX) that are updated independently using the DQN loss function \cite{mnih_human-level_2015}, but with respect to $Q_i$. In ICL each agent adjusts the reward in the loss for the network update, as described in Eq. \ref{eq:theo_eq2}. Note that, in this paper, Independent Deep Q-learning (IDQL) refers to the use of independent DQNs for each agent in MARL, similarly to as introduced in \cite{tampuu_multiagent_2015}.

\subsection{Causality Effect in the Environments Used}\label{sec:caus_envs}
\textbf{Predator-Prey} A team of 4 agents needs to capture 2 moving preys. Every time a prey is caught, the team receives a positive reward. However, since two agents at the same time are enough to capture a prey, some of them might not have been involved in the capture and can become lazy. To adjust the credit, we use a causality concept to relate the individual observations with the team reward: at each timestep $t$, each agent will only be rewarded if there is a positive reward (capture), and there is at least one prey in the observation mask of the agent in the moment before the capture.

\textbf{Lumberjacks} A team of 4 agents has to chop all the existing trees in the map. Each tree has a randomly assigned level $l$, where $l$ agents are needed at the same time to cut the tree. There is a team reward every time a tree is cut, but the credit to the agents is adjusted under the conditions: there is a positive reward (tree is cut) and 1) there is a tree in the observation mask of the agent in the moment before the tree was cut, and 2) the number of agents seen by the agent in its observation mask (including itself) is greater or equal than the level of at least one of the trees in the observation mask of the agent.

\begin{figure}[t!]
    \centering
    \includegraphics[width=0.7\columnwidth]{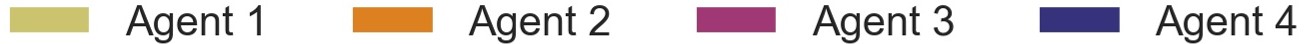}
    \\
    \vspace{0.00mm} 
    \subfigure[]{\label{fig:add_res_d}\includegraphics[width=0.45\columnwidth]{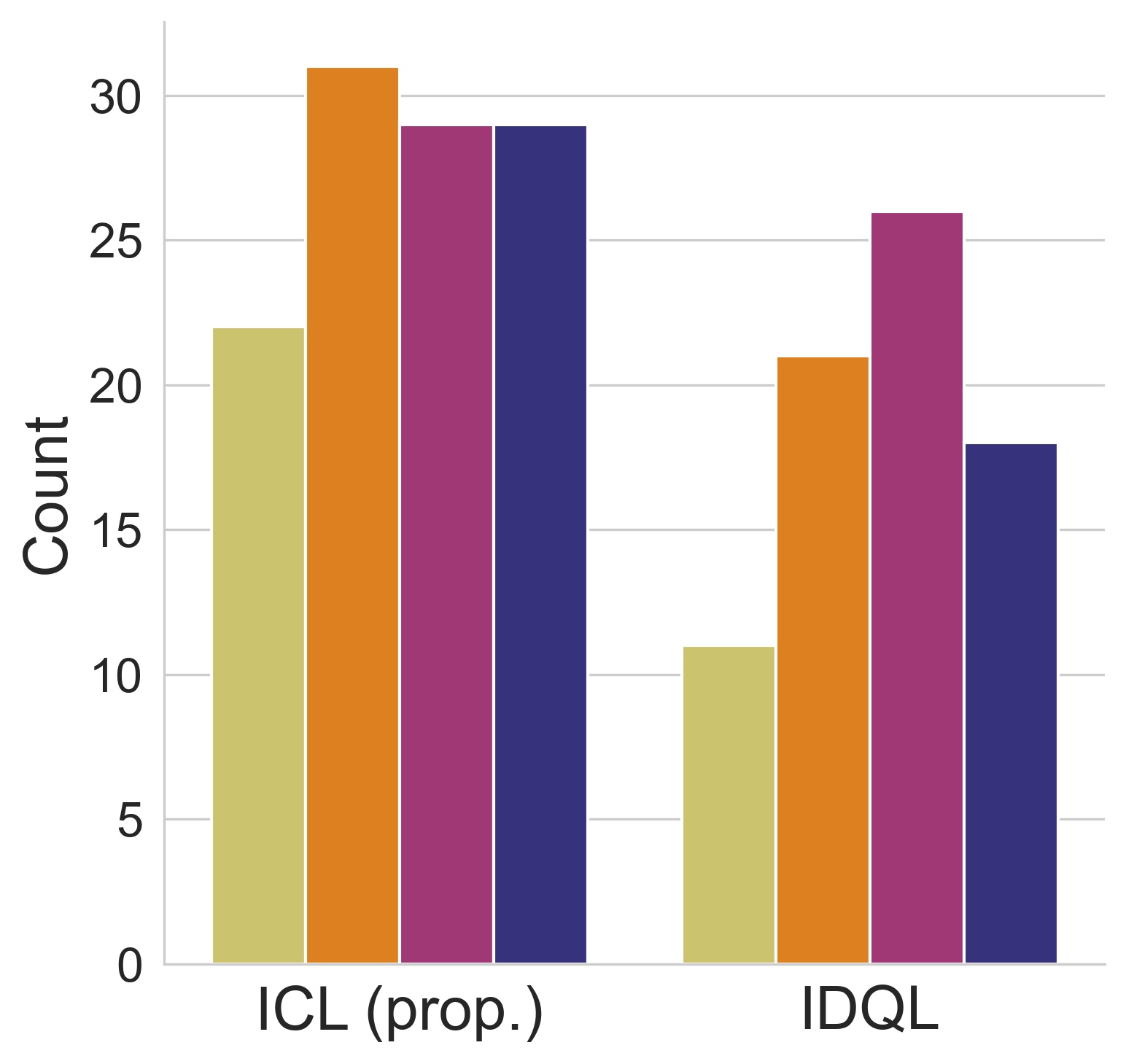}}
    \subfigure[]{\label{fig:add_res_e}\includegraphics[width=0.45\columnwidth]{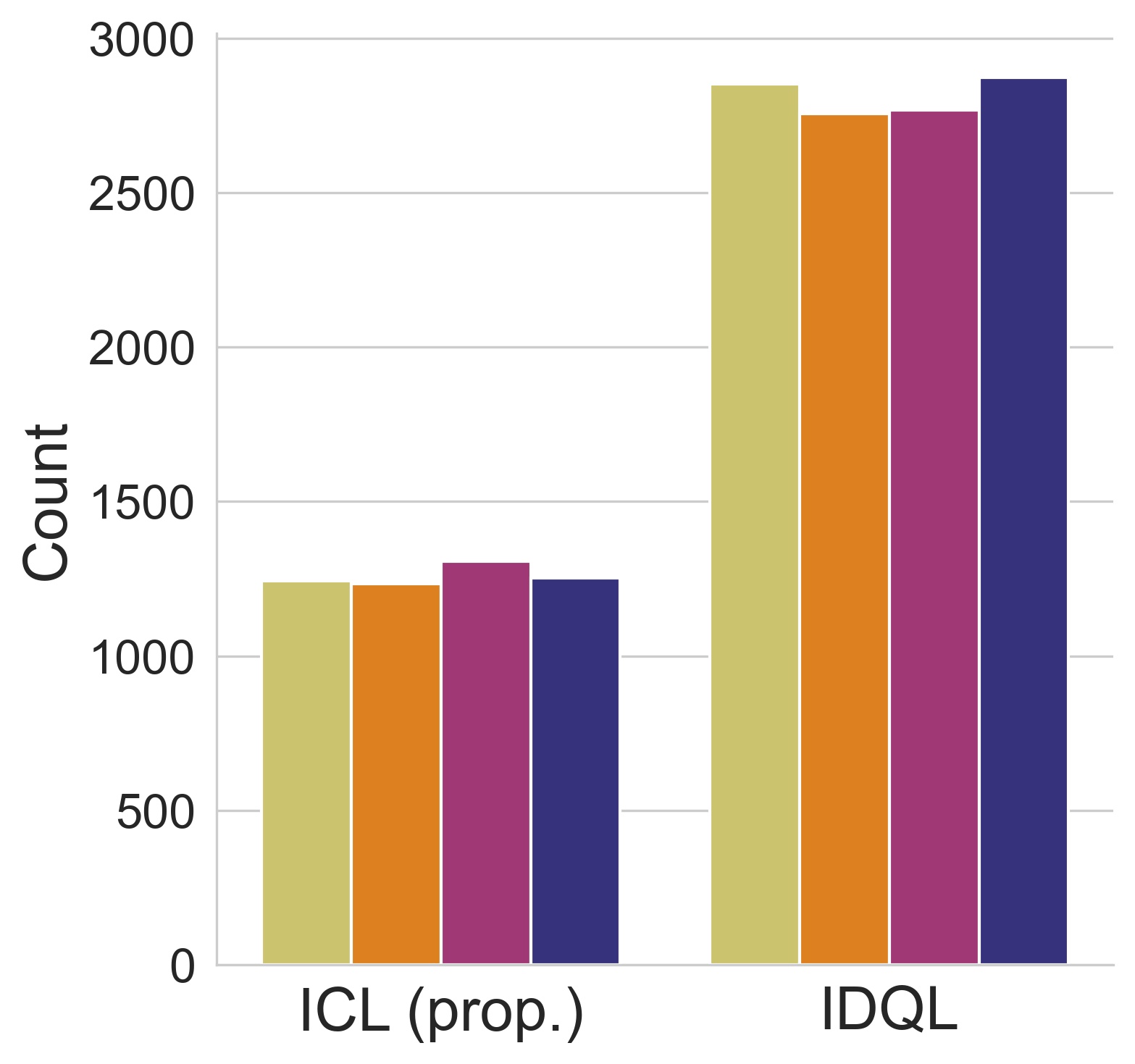}}
    \caption{Behaviour of trained agents for ICL vs IDQL. (a) preys caught per agent in Predator-Prey (50 episodes); (b) total distance from each agent to the other agents over one trained episode of Lumberjacks (if they move closer they will solve the task more easily).}
    \label{fig:add_results}
\end{figure}

\subsection{Amortized Causal Discovery for MARL}
We introduce a second approach ACD-MARL, intending to support that causal relations exist in the dynamics of MARL and can be inferred. To this end, we show how ACD can be combined with MARL problems and used to infer observation-reward causal relations within teams of agents. ACD is a deep learning causality detection method and is theoretically aligned with Granger Causality \cite{lowe_amortized_2022}. ACD models causal relations among a set of input series using an encoder-decoder architecture where the latent space represents the edges of a causal graph. For the proposed, method we use a modified ACD architecture applied to MARL problems. The goal is to detect causal relations between individual observations and the team reward. We start by gathering MARL samples that are given to the model, where the resulting latent space denotes a $(N+1)\times (N+1)$ adjacency matrix that represents the edges of the learned causal graph across the $N+1$ MARL time series. From this graph, we extract the learned values that correspond to $c_i(o_i,r)$ in the ICL ground truth.

\section{Experiments and Results}
We start by evaluating the performance of ICL, using QMIX and IDQL as benchmarks. In Fig. \ref{fig:res_a} we can see that all methods learned policies that solve Predator-Prey, but, while ICL learns the task sooner, IDQL stays slightly below the other methods. Fig. \ref{fig:res_c} shows that QMIX can easily solve Lumberjacks, demonstrating the usefulness of centralised training in this scenario. Although ICL does not converge as quickly as QMIX in this case, it eventually reaches the same level, while IDQL fails to achieve an optimal reward. When we look at the behaviours learned by IDQL and ICL agents, Fig. \ref{fig:add_res_d} and \ref{fig:add_res_e} show that IDQL agents do not cooperate nearly as much as ICL agents. In both tasks, there is a higher participation of the agents towards helping the team when they use a causality factor. This also explains the better performances as a team in Fig. \ref{fig:res_a} and \ref{fig:res_c}.

To evaluate the second method ACD-MARL, we use the causality detection mechanisms of ICL presented in the previous section as the ground truth. The results show that the predicted $o\rightarrow r$ causal relations (that correspond to $c_i$) by ACD-MARL are close to the ground truth relations of ICL, enforcing that causal relations are indeed present in the underlying dynamics of MARL. The method achieves 68\% and 67\% of accuracy for Predator-Prey and Lumberjacks, respectively. It is important to note that the cases where ACD-MARL fails to predict correspond mostly to false positive cases (29\%/32\% and 28\%/33\%). The false positive cases are not as harmful for the performance of the team, since false positives boil down to simple independent learning when we use the predicted $c_i$ to adjust the credit assignment. By having these sparsely together with a high number of correct credit assignment causal predictions, it would still result in an improvement of team performance and individual behaviours.

\begin{acks}
This work was funded by the Engineering and Physical Sciences Research Council in the United Kingdom (EPSRC), under the grant number EP/T000783/1.
\end{acks}





\end{document}